# OPENMV: A PYTHON POWERED, EXTENSIBLE MACHINE VISION CAMERA

Ibrahim Abdelkader[1], Yasser El-Sonbaty[1] and Mohamed El-Habrouk[2]
[1]*Dept. of Computer Science, Arab Academy for Science & Technology, Alexandria, Egypt*
[2]*Dept. of Electrical Engineering, Faculty of Engineering, Alexandria, Egypt*

**ABSTRACT**

Advances in semiconductor manufacturing processes and large scale integration keep pushing demanding applications further away from centralized processing, and closer to the edges of the network (i.e. Edge Computing). It has become possible to perform complex in-network image processing using low-power embedded smart cameras, enabling a multitude of new collaborative image processing applications. This paper introduces OpenMV, a new low-power smart camera that lends itself naturally to wireless sensor networks and machine vision applications. The uniqueness of this platform lies in running an embedded Python3 interpreter, allowing its peripherals and machine vision library to be scripted in Python. In addition, its hardware is extensible via modules that augment the platform with new capabilities, such as thermal imaging and networking modules.

## 1. INTRODUCTION

Embedded smart cameras have a wide range of applications from automation and robotics to complex in-network image processing such as distributed object tracking and localization (Ercan et al, 2007) and distributed surveillance (Bramberger et al, 2006). This paper presents OpenMV—a new low-cost, low-power embedded smart camera platform for machine vision and wireless sensor networks applications. OpenMV is designed with low-cost and usability in mind. It can be scripted in Python 3 and comes with an extensive machine vision library, an IDE and example scripts. The IDE allows viewing the frame-buffer and uploading and executing scripts via serial over USB or sockets, enabling remote development and deployment.

The remainder of this paper is organized as follows: Section 2 presents a survey of related low-power smart camera platforms. Section 3 and 4 introduce the proposed smart camera hardware and software architectures. Finally, the conclusions and future work are presented in Section 5 as well as a comparison between existing platforms and the proposed platform.

## 2. RELATED WORK

This chapter presents a survey of existing low-power embedded smart camera platforms. Their design specifications, main features and limitations are reviewed. In addition, a comparison between the following platforms and the proposed platform is presented in Table 1 at the end Section 5.

### 2.1 eCam

eCam (Park & Chou, 2006) is an ultra compact (1.1x0.7 inches) wireless image sensor node that consumes about 230mW of power. eCam combines a single package VGA sensor/lens with an Eco (Park & Chou, 2006) wireless sensor platform and a 170mAh Lithium-Polymer battery. See Figure 1. eCam supports JPEG compression via an external JPEG codec, which also serializes the image data to the host controller (located on the Eco node) due to the lack of a hardware camera interface. The Eco node's MCU (nRF24E1) is a VLSI chip integrating a radio transceiver and an 8051 core that runs at a few MHz with a few KiBs of RAM. The radio transceiver has a maximum data-rate of 1Mbps which can only support 1.5FPS streaming.

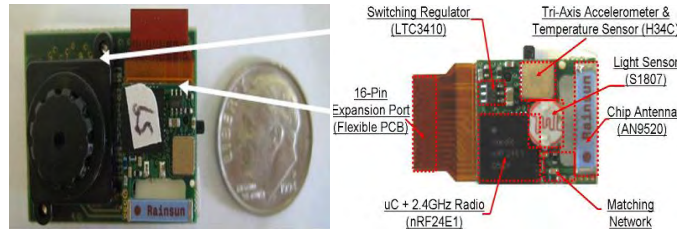

Figure 1. (a) eCam: Ultra compact camera node. (b) Echo Node

## 2.2 CMUCam3

CMUCam3 (Rowe et al, 2007) is an embedded camera based on NXP's LPC2106 MCU (a 32-bit 60MHz ARM7 MCU with on-chip 64KiB of RAM and 128KiB of flash) and an OV6620 image sensor, see Figure 2.

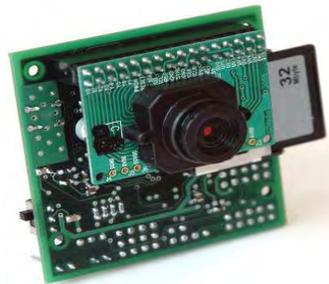

Figure 2. CMUcam3: Main board, sensor and an MMC memory card for mass storage

Due to its limited memory, the CMUCam3 stores frames in a 1MiBs FIFO chip (AL4V8M440). This approach increases cost and power requirements. Additionally, it does not allow the image to be accessed randomly without copying parts of the frame to the SRAM first, which slows down the frame-rate. The CMUCam3 implements a set of optimized image processing algorithms in C, in addition to image compression libraries and a scripting engine. CMUCam3 consumes about 500mW of power.

## 2.3 Pixy (CMUCam5)

Pixy (CMUcam5, 2015) is the new version of CMUCAM3, featuring a 200MHz dual core (ARM Cortex-M4/M0) LPC4330 MCU with 264KiBs of RAM 1MiB of Flash, see Figure 3.

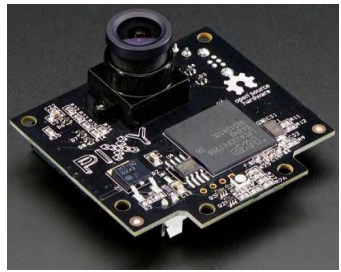

Figure 3. Pixy (CMUCam5)

Unlike its predecessor, Pixy is designed specifically to recognize colored objects. Pixy does not provide a scripting language, nor does it provide any other image processing features. However, the default firmware image can be replaced with a custom one. For extensions, the Pixy camera uses an I/O header on the back side to breakout the UART, SPI and I2C peripherals, possibly to be used for interfacing radio and/or hardware modules.

### 2.4 Mesheye

MeshEye (Hengstler et al, 2007) is a smart stereo-camera targeted at intelligent surveillance applications. MeshEye is based on a 50MHz ARM7 (AT91SAM7S) processor, with 64KiB of RAM and 256 KiB of flash, see Figure 4. MeshEye has a unique stereo-vision system that continuously tracks the position, range, and size of moving objects. This information is used to trigger a higher resolution color camera to capture an image of the object for further processing.

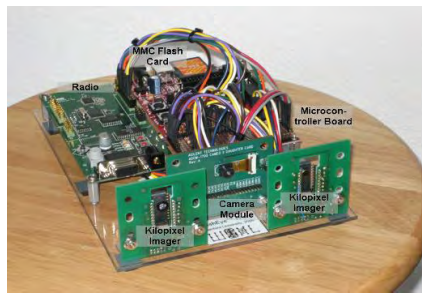

Figure 4. MeshEye Prototype

### 2.5 MicrelEye

MicrelEye (Kerhet et al, 2007) is a fully integrated wireless camera, designed for cooperative distributed image processing applications. MicrelEye consists of an Atmel MCU+FPGA system-on-chip (SoC), an OV7620 VGA image sensor and an LMX9820A Bluetooth transceiver, see Figure 5.

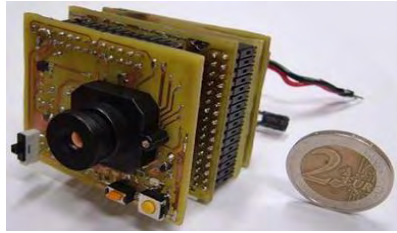

Figure 5. MicrelEye node

MicrelEye nodeThe MicrelEye architecture is unique in that it uses a SoC with configurable logic blocks. Using FPGAs allows exploiting parallelism to optimize some image processing algorithms. For example, optical flow performance can be optimized using FPGAs (Nagy et al, 2006). In addition, an optimized support vector machine-like (SVM-like) algorithm is implemented on MicrelEye for people detection and counting. The algorithm is partially implemented on the FPGA. The first steps (image readout, background subtraction...) are implemented on FPGA, whereas feature extraction and SVM are implemented on the microcontroller. In addition to image processing, the FPGA is also used to implement the image sensor interface and to access an external 1MBs SRAM, which is used to store frames for later processing.

## 3. OPENMV SMART CAMERA

In this section, the OpenMV platform is introduced and discussed in detail, including the image sensor, microcontroller, networking, extensions, software and firmware.

### 3.1 Overview

The main board measures 1.4"x1.2" and consists of the image sensor, the MCU, power supply, a micro-SD card slot and extension headers, see Figure 6. In addition, built-in RGB and IR LEDs are provided on board, for status indication and night vision, respectively.

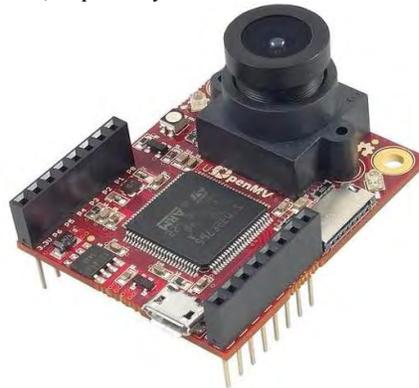

Figure 6. OpenMV's main board

### 3.2 Image Sensor

OpenMV uses OmniVision's OV7725 VGA CMOS sensor. The OV7725 supports cropping and windowing to output arbitrary frame sizes, as well as standard resolutions. The OV7725 was chosen mainly for its low-cost and its high sensitivity in low-light operation (enabled by its 6.0x6.0μm pixels). The OV7725 does support JPEG, however the MCU has a hardware JPEG encoder. JPEG compression is mainly used to transfer frames to the host for debugging, however it can enable advanced features such as medical images compression (El-Sonbaty et al, 2003).

### 3.3 Microcontroller

OpenMV is based on the STM32F7 ARM Cortex-M7 dual-issue MCU running at 216MHz. The MCU features 512KiBs SRAM, 2MiBs flash (part of which is used for lookup tables and internal flash filesystem) DMA and DMA2D, USB OTG, a single precision FPU, DSP instructions, a digital camera interface (DCMI), JPEG encoder, timers and multiple serial peripheral interfaces such as I2C, SPI and UART.

The MCU's SRAM is divided into two non-contiguous blocks; a main block and a core-coupled memory (CCM). The main block is used for the frame buffer, and for storing computed integral images (Crow, 1984) and other temporary images when needed, while the CCM is used exclusively for the stack, heap and data.

### 3.4 Networking

The main OpenMV board is decoupled from the networking modules using extension modules (or shields). This decoupling allows the camera to work with different networks (such as WiFi, BLE or Zigbee) using different radio transceivers to easily integrate the sensor into existing network infrastructures.

### 3.5 Power Consumption

To minimize idle power consumption, the CPU can enter low-power modes from which it can be awaken via interrupts. Additionally, in a few key places in the code, such as waiting for an image readout, the *WFI* (Wait For Interrupt) instruction is executed to force the processor to suspend execution until an IRQ is received. Furthermore, the MCU supports switching to lower frequencies (frequency scaling).

### 3.6 Extensions

The extension headers breakout the ADC/DAC, PWM, I2C, SPI and UART interfaces, allowing the camera to be interfaced to motors, network modules and other sensors such as thermal and distance sensors. A low-resolution thermal imaging shield was designed allowing the sensor to be used for presence detection. Other shields including WiFi, BLE and LCD shields were also designed, see Figure 7.

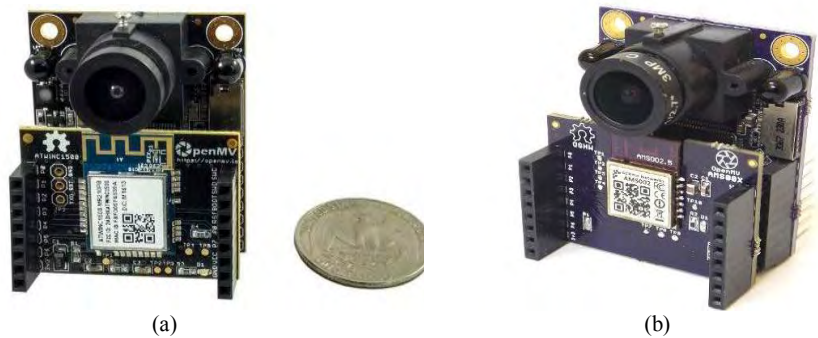

(a)          (b)

Figure 7. OpenMV hardware extensions. (a) WiFi module. (b) BLE (Bluetooth Low-Energy) module

## 4. OPENMV SOFTWARE ARCHITECTURE

The software architecture consists of a number of hardware abstraction layers (HALs), middleware and userspace software, see Figure 8.

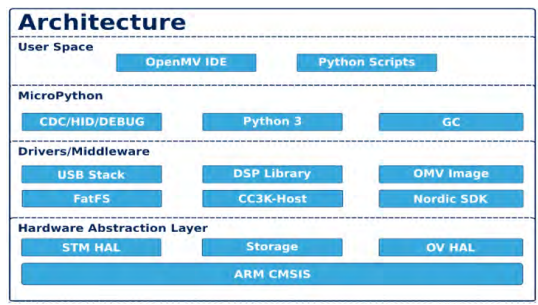

Figure 8. Software architecture shows the hardware abstraction layers, middleware and userspace software

The glueware/middleware layer includes the USB stack, FatFS (Cha, 2005) filesystem, DSP libraries, network drivers and optimized machine vision code. On top of those layers, the camera runs a lightweight implementation of Python3 called MicroPython (MicroPython, 2013) which allows user access to peripheral drivers and compiled machine vision code using Python scripts. The userspace level consists of a cross-platform IDE designed specifically for the camera, which can view the frame buffer, access sensor controls, upload scripts and run them on the camera via USB (or WiFi/BLE if available).

### 4.1 Machine Vision Support

The OpenMV image processing library implements a fairly comprehensive set of image processing algorithms and supports multiple image formats (PGM/PPM, BMP, JPEG for still images, and MJPEG and GIFs for videos). The image processing library is implemented in optimized C code while exporting a user-friendly API which can be called from Python scripts.

### 4.2 Basic Image Processing Functions

The image processing library includes basic image handling functions such as loading/saving images from the filesystem to memory (or the frame buffer), cropping, scaling and blending. In addition, basic drawing functions for visualizing results are also implemented, such as drawing lines, rectangles, circles and strings, and setting/getting image information and pixels. Furthermore, the library provides image filtering functions are implemented such as median, midpoint, Gaussian smoothing functions and histogram equalization.

### 4.4 Advanced Image Processing Functions

The OpenMV image processing library also provides several advanced image processing functions, such as QR Code detection and decoding, ApriTags (Olson, 2011) support, face and eyes detection using Viola-Jones Haar cascade (Viola & Jones, 2001) and iris estimation by analyzing the gradient vectors in an eye (Timm & Barth, 2011). See Figure 9.

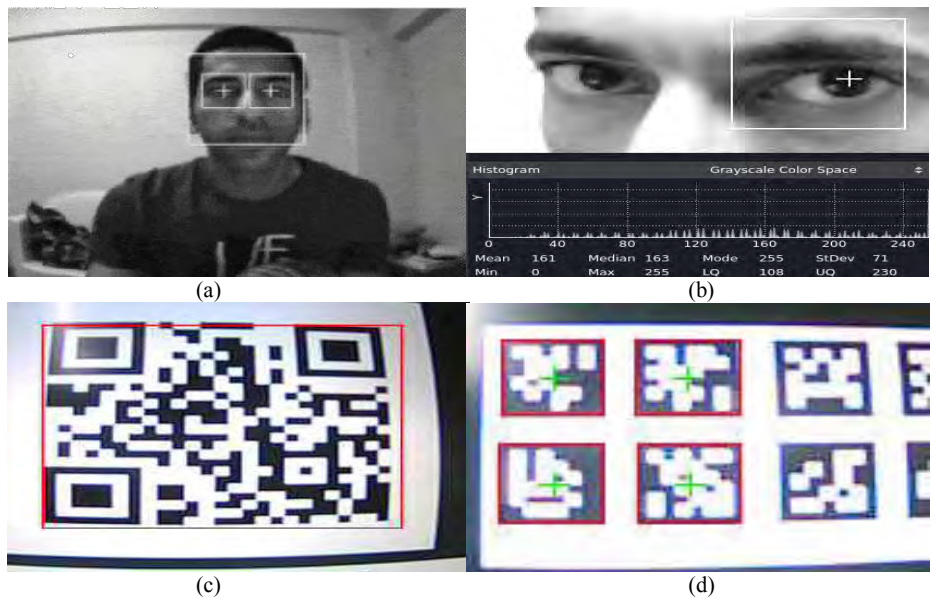

Figure 9. Advanced machine vision features: (a) Face and eyes detection running on OpenMV
(b) Iris localization

Many algorithms have been proposed for matching partially occluded objects, such as matching connected lines (El-Sonbaty & Ismail, 2003). To accomplish this, the image library includes FAST (Rosten & Drummond, 2006)-ORB (Rublee et al, 2011) detector for keypoint extraction and description respectively, in addition to Histogram of Oriented Gradients (HoG) (Dalal & Triggs, 2005; Bilinski et al, 2009). Additionally, the library implements fast template matching via Normalized Cross Correlation (NCC) (Lewis, 1995) with both extensive search and Diamond Search (DS) (Zhu & Ma, 2000), optical flow for estimating the direction of movement and finally, the Canny edge detector and Hough transform for detection of lines in an image.

## 5. CONCLUSION

This paper presented OpenMV, a machine vision platform, suitable for use in wireless sensor networks and the IoT. The platform runs an embedded Python3 interpreter and implements an optimized image processing library which can be called from Python user scripts. The OpenMV main board is decoupled from networking modules to allow interfacing to different networks. In addition to its commercial availability, OpenMV is open-source hardware and software, which means it can be built from source for use in research purposes. Future work includes hardware upgrade to the new STM32FH 1-1.5MiBs SRAM/400MHz, Cortex-M7 MCU.

Table 1. A comparison of image sensor platforms

| Sensor | eCam | CMUCam3 | Pixy (CMUCam5) | MeshEye | MicrelEye | OpenMV |
|---|---|---|---|---|---|---|
| Power | 230mW | 500mW | 700mW | 260mW | N/A | 500mW |
| Networking | nRF24E1 | N/A | N/A | Zigbee | BLE | WiFi/BLE/Zigbee/nRF24x |
| Processor Type | VLSI uC/2.4Radio | µC | µC | µC | µC+FPGA | µC |
| Clock Speed | 20MHz | 60MHz | 204MHz | 55MHz | 14MHz | 216/400MHz (Dual issue) |
| RAM | 4KiB SRAM | 64KiB SRAM | 264KiB SRAM | 64KiB SRAM | 16KiB SRAM | 512KiB/1.5MiBs SRAM |
| ROM | External EEPROM | 128KiB | 1MiB | 256KiB | 20KiB | 1-2MiB |
| External Storage | No | SD Card | No | MMC | No | uSD/Flash |
| FPU/DSP Or Acceleration | No | No | FPU, DSP | No | FPGA | FPU, DSP |
| Image Buffer | N/A | FIFO | Internal SRAM | MMC | External SRAM | Internal SRAM |
| Programmable/Scriptable | No | Yes | No | No | No | MicroPython |
| Image Sensor | OV7640 (0.3MP) | OV6620 (0.1MP) | OV9715 (1MP) | ADNS-3060 ADCM-2700 | OV7640 (0.3MP) | OV7725 (0.3MP) |
| Hardware JPEG Encoder | (External) Yes | No | No | No | No | Yes |
| Image Processing | No | Face detection, Convolutions, color tracking, Frame Differencing | Color Tracking | Object Detection and Tracking | Image Classification. | Face detection FAST/ORB Iris/Eye , Optical flow, Color tracking, QR-Code, AprilTags |
| Open Source | No | Yes | Yes | No | No | Yes |
| BOM Cost | N/A | N/A | N/A | N/A | N/A | $16@1000s |
| Commercially available | No | No | Yes | No | No | Yes |
| Dimensions | 2.0x2.8 cm | 5.5x5.7 cm | 5.3x5.0 cm | N/A | 5x5x3 cm$^3$ | 3.0x3.5 cm |